\documentclass[10pt,twocolumn,letterpaper]{article}
\pdfoutput=1
\newcommand*{\affaddr}[1]{#1} 
\newcommand*{\affmark}[1][*]{\textsuperscript{#1}}
\newcommand*{\email}[1]{\texttt{\small#1}}
\usepackage{iccv}
\usepackage{times}
\usepackage{epsfig}
\usepackage{graphicx}
\usepackage{amsmath}
\usepackage{amssymb}
\usepackage[T1]{fontenc} 
\usepackage{subfig,graphicx}
\usepackage{booktabs}
\usepackage{multirow}
\usepackage{algorithm}
\usepackage{algorithmic} 
\usepackage{caption}
\usepackage[pagebackref=true,breaklinks=true,letterpaper=true,colorlinks,bookmarks=false]{hyperref}

\iccvfinalcopy 
\usepackage{bbm}
\def\iccvPaperID{5415} 

\ificcvfinal\fi

\begin{document}

\title{Less is More: Focus Attention for Efficient DETR}

\author{
	Dehua Zheng\affmark[1,2] \quad Wenhui Dong\affmark[2] \quad Hailin Hu\affmark[2] \quad Xinghao Chen\affmark[2]\quad Yunhe Wang\affmark[2]\footnotemark[1]\\
	\affaddr{\affmark[1]Huazhong University of Science and Technology\quad \affmark[2]Huawei Noah’s Ark Lab\\ }\\
	\email{dwardzheng@hust.edu.cn\quad \{wenhui.dong, hailin.hu, xinghao.chen, yunhe.wang\}@huawei.com\quad}
}

\maketitle

\renewcommand{\thefootnote}{\fnsymbol{footnote}}
\footnotetext[1]{Corresponding author} 
\ificcvfinal\thispagestyle{empty}\fi

\begin{abstract}
DETR-like models have significantly boosted the performance of detectors and even outperformed classical convolutional models. However, all tokens are treated equally without discrimination brings a redundant computational burden  in the traditional encoder structure. The recent sparsification strategies exploit a subset of informative tokens to reduce attention
complexity maintaining performance through the sparse encoder. But these methods tend to rely on unreliable model statistics. Moreover, simply reducing the token population hinders the detection performance to a large extent, limiting the application of these sparse models. We propose Focus-DETR, which focuses attention on more informative tokens for a better trade-off between computation efficiency and model accuracy. Specifically, we reconstruct the encoder with dual attention, which includes a token scoring mechanism that considers both localization and category semantic information of the objects from multi-scale feature maps. We efficiently abandon the background queries and enhance the semantic interaction of the fine-grained object queries based on the scores. Compared with the state-of-the-art sparse DETR-like detectors under the same setting, our Focus-DETR gets comparable complexity while achieving 50.4AP~(\textbf{+2.2}) on COCO. The
code is available at  \href{https://github.com/huawei-noah/noah-research}{torch-version} and \href{https://gitee.com/mindspore/models/tree/master/research/cv/Focus-DETR}{mindspore-version}.
\end{abstract}

\section{Introduction}

\begin{figure}[t]
	\begin{center}
	\includegraphics[width=1.0\linewidth]{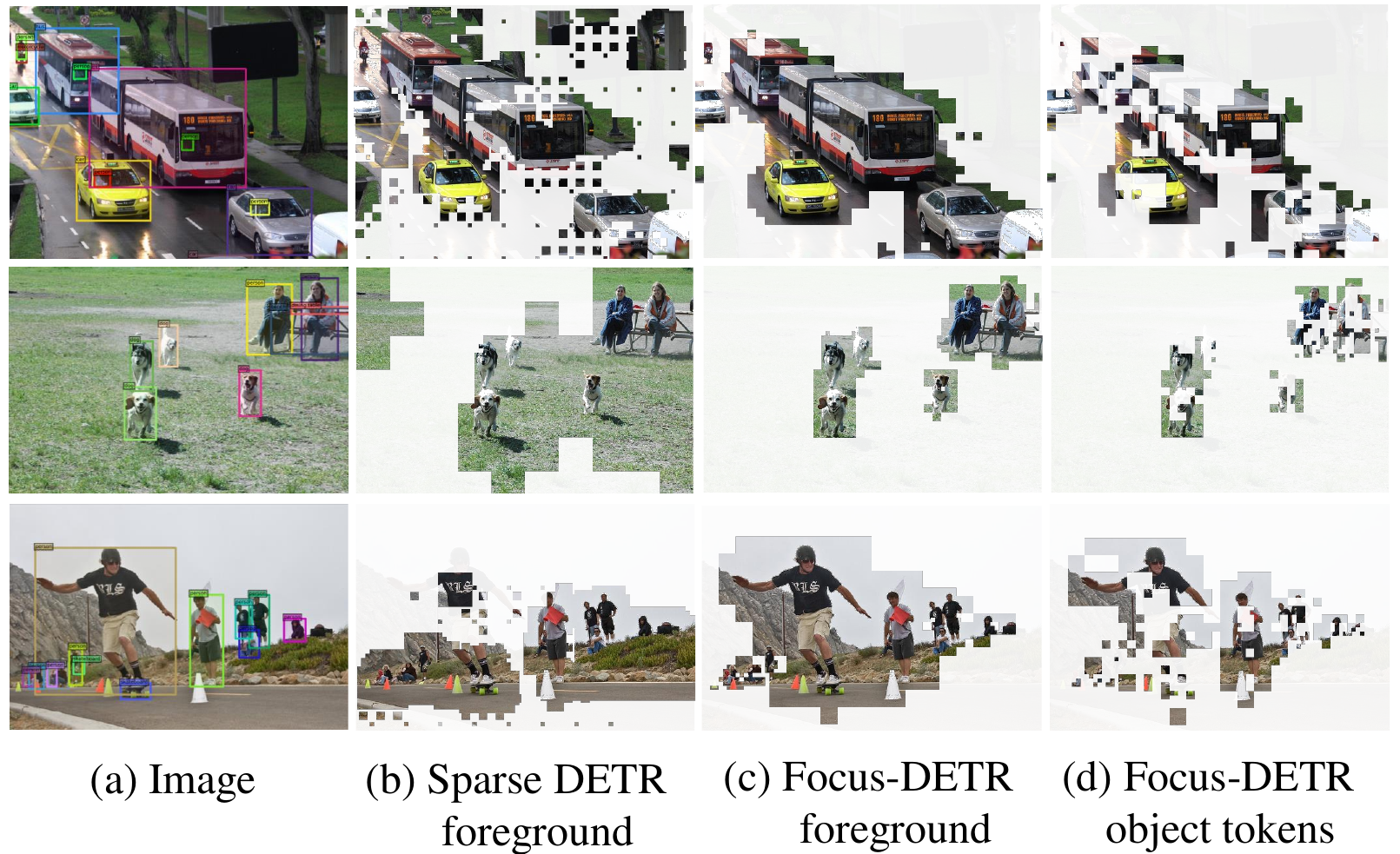}
	\end{center}
	\vspace{-0.7em}
	\caption{Visualization and comparison of tokens selected by Sparse DETR~\cite{sparse_DETR} and our Focus-DETR. (a) is the original images, (b) and (c) represent the foreground selected by models. (d) indicates the object tokens with more fine-grained category semantic. Patches with smaller sizes come from higher-level features.
	}
	\label{fig:patchvisual}

\end{figure}

Object detection is a fundamental task in computer vision that aims to predict the bounding boxes and classes of objects in an image, as shown in Fig.~\ref{fig:patchvisual}~(a), which is of great importance in real-world applications. DETR proposed by Carion \textit{et al.}\cite{DETR} uses learnable queries to probe image features from the output of Transformer encoders and bipartite graph matching to perform set-based box prediction. DETR-like models~\cite{DAB_DETR,DINO,DN_DETR,efficient_DETR,conditional_DETR,sparse_DETR,group_DETR,anchor_DETR,Deformable_DETR} have made remarkable progress and gradually bridged the gap with the detectors based on convolutional neural networks.

\begin{figure}[t]
	\begin{center}
		\includegraphics[width=1.0\linewidth]{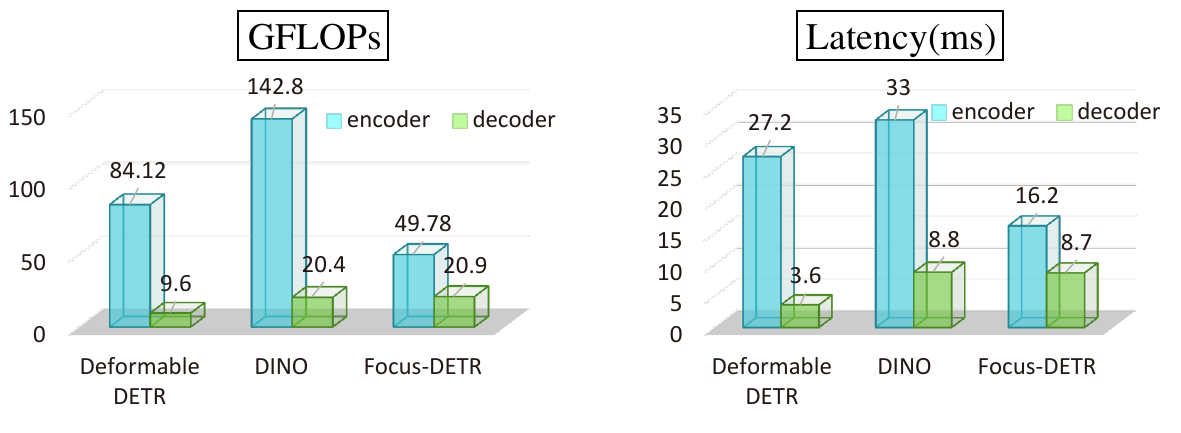}
	\end{center}
\vspace{-1.3em}
	\caption{Distribution of calculation cost and latency in the Transformer part of the DETR-like models, e.g., Deformable DETR~\cite{Deformable_DETR}, DINO~\cite{DINO} and our Focus-DETR.} 
	\label{fig:flops}
	\vspace{-0.8em}
\end{figure}

Global attention in the DETR improves the detection performance but suffers from computational burden and inefficiency due to redundant calculation without explicit discrimination for all tokens. To tackle this issue, Deformable DETR~\cite{Deformable_DETR} reduces the quadratic complexity to linear complexity through key sparsification, and it has developed into a mainstream paradigm due to the advantages of leveraging multi-scale features.  Herein, we further analyze the computational burden and latency of components in these models~(Fig.~\ref{fig:flops}). As shown in Fig.~\ref{fig:flops}, we observe that the calculation cost of the encoder is 8.8$\times$ that of the decoder in Deformable DETR~\cite{Deformable_DETR} and 7.0$\times$ in DINO~\cite{DINO}. In addition, the latency of the encoder is approximately 4$ \sim $8 times that of the decoder in Deformable DETR and DINO, which emphasizes the necessity to improve the efficiency in the encoder module. In line with this, previous works have generally discussed the feasibility of compressing tokens in the transformer encoder. For instance, PnP-DETR~\cite{pnp_DETR} abstracts the whole features into fine foreground object feature vectors and a small number of coarse background contextual feature vectors. IMFA~\cite{IFMA} searches key points based on the prediction of decoder layer to sample multi-scale features and aggregates sampled features with single-scale features. Sparse DETR~\cite{sparse_DETR} proposes to preserve the 2D spatial structure of the tokens through query sparsity, which makes it applicable to Deformable DETR~\cite{Deformable_DETR} to utilize multi-scale features. By leveraging the cross-attention map in the decoder as the token importance score, Sparse DETR achieves performance comparable to Deformable DETR only using 30\% of queries in the encoder.

Despite all the progress, the current models~\cite{pnp_DETR,sparse_DETR} are still challenged by sub-optimal token selection strategy. As shown in Fig.~\ref{fig:patchvisual} (b), the selected tokens contain a lot of noise and some necessary object tokens are obviously overlooked. In particular, Sparse DETR's supervision of the foreground predictor relies heavily on the decoder's cross-attention map~(DAM), which is calculated based on the decoder's queries entirely from encoder priors. Preliminary experiments show severe performance decay when the Sparse DETR is embedded into the models using learnable queries due to weak correlation between DAM and the retained foreground tokens. However, state-of-the-art DETR-like models, such as DINO~\cite{DINO}, have proven that the selected features are preliminary content features without further refinement and could be ambiguous and misleading to the decoder. In this case, DAM's supervision is inefficient. Moreover, in this monotonous sparse encoder, the number of retained foreground tokens remains numerous, and performing the query interaction without more fine-grained selection is not feasible due to computational cost limitations.


To address these issues, we propose Focus-DETR to allocate attention to more informative tokens by stacking the localization and category semantic information. Firstly, we design a scoring mechanism to determine the semantic level of tokens. \underline{\textbf{F}}oreground \underline{\textbf{T}}oken \underline{\textbf{S}}elector~(FTS) aims to abandon background tokens based on top-down score modulations across multi-scale features. We assign \{1,0\} labels to all tokens from the backbone with reference to the ground truth and predict the foreground probability. The score of the higher-level tokens from multi-scale feature maps modulates the lower-level ones to impose the validity of selection. To introduce semantic information into the token selection process, we design a multi-category score predictor. The foreground and category scores will jointly determine the more fine-grained tokens  with strong category semantics, as shown in Fig.~\ref{fig:patchvisual}~(d). Based on the reliable scores and selection from different semantic levels, we feed foreground tokens and more fine-grained object tokens to the encoder with dual attention. Thus, the limitation of deformable attention in distant information mixing is remedied, and then the semantic information of foreground queries is enhanced by fine-grained token updates.

To sum up, Focus-DETR reconstructs the encoder's calculation process with dual attention based on obtaining more accurate foreground information and focusing on fine-grained tokens by gradually introducing semantic information, and further enhances fine-grained tokens with minimal calculation cost. Extensive experiments validate Focus-DETR's performance. Furthermore, Focus-DETR is general for DETR-like models that use different query construction strategies. For example, our method can achieve 50.4AP (\textbf{+2.2}) on COCO compared to Sparse DETR with a similar computation cost under the same setting.

\section{Related work}
\textbf{Transformer-based detectors.} Recently, Carion \textit{et al}.\cite{DETR} proposed an end-to-end object detector named DETR (DEtection TRansformer) based on Vision Transformer~\cite{Vit}. DETR transforms object detection into a set prediction task through the backbone, encoder, and decoder and supervises the training process through Hungarian matching algorithms. A lot of recent works\cite{DAB_DETR,DN_DETR,Deformable_DETR,DINO,conditional_DETR,conditional_DETR_v2,SAMDETR,group_DETR,DynamicDETR} have boosted the performance of Transformer-based detectors from the perspective of accelerating training convergence and improving detection precision. Representatively DINO\cite{DINO} establishes DETR-like models as a mainstream detection framework, not only for its novel end-to-end detection optimization, but also for its superior performance.
Fang \textit{et al}. \cite{YOLOS} propose YOLOS and reveal that object detection can be accomplished in a pure sequence-to-sequence manner with minimal additional inductive biases. Li \textit{et al}.\cite{VITDET} propose ViTDet to explore the plain, non-hierarchical ViT as a backbone for object detection. Dai \textit{et al}.\cite{upDETR} propose a pretext task named random query patch detection to Unsupervisedly Pre-train DETR (UP-DETR) for object detection. IA-RED$^2$~\cite{IA-RED} introduces an interpretable module for dynamically discarding redundant patches. 

\textbf{Lightweight Vision Transformers.} 
As we all know, vision Transformer~(ViT) suffers from its high calculation complexity and memory cost. Lu \textit{et al}.~\cite{DynamicViT} propose an efficient ViT with dynamic sparse tokens to accelerate the inference process. Yin \textit{et al}.\cite{AViT} adaptively adjust the inference cost of ViT according to the complexity of different input images. Xu \textit{et al}.\cite{Evo-vit} propose a structure-preserving token selection strategy and a dual-stream token update strategy to significantly improve model performance without changing the network structure. Tang \textit{et al}.~\cite{PatchSlimming} presents a top-down layer by layer patch slimming algorithm to reduce the computational cost in pre-trained Vision Transformers. The core strategy of these algorithms and other similar works\cite{LearnedToken, SPViT,AdaptiveSparseViT} is to abandon redundant tokens to reduce the computational complexity of the model.

\begin{figure*}
	\begin{center}
		\includegraphics[width=14cm]{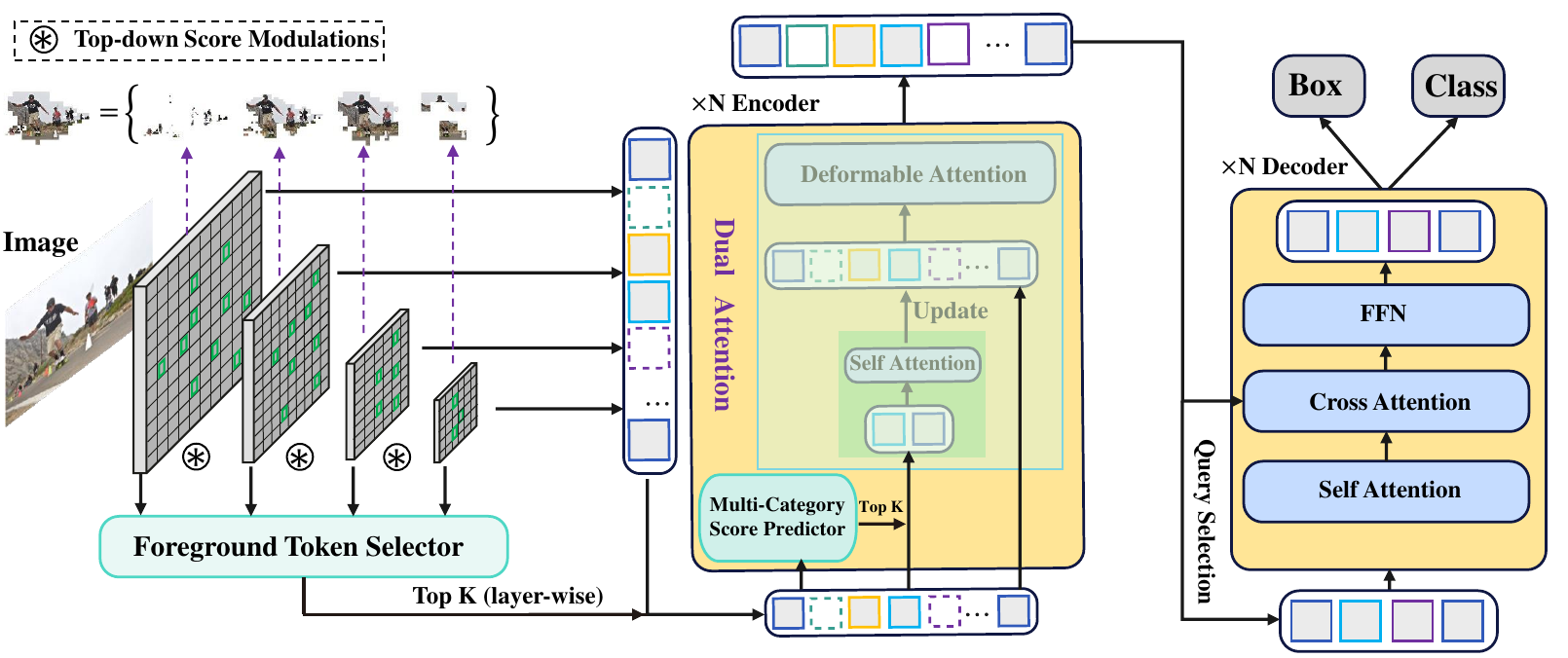}
	\end{center}
	\vspace{-1.0em}
	\caption{The architecture overview of the proposed Focus-DETR. Our Focus-DETR comprises a backbone network, a Transformer encoder, and a Transformer decoder. We design a foreground token selector~(FTS) based on top-down score modulations across multi-scale features.  And the selected tokens by a multi-category score predictor and foreground tokens go through the encoder with dual attention to remedy the limitation of deformable attention in distant information mixing.}
	\label{fig:overview}
	\vspace{-0.8em}
\end{figure*}

In addition to the above models focused on sparsity backbone structure applied on classification tasks, some works\cite{sparse_DETR, pnp_DETR} lie in reducing the redundant calculation in DETR-like models. Efficient DETR~\cite{efficient_DETR} reduces the number of layers of the encoder and decoder by optimizing the structure while keeping the performance unchanged. PnP-DETR and Sparse DETR have achieved performance comparable to DETR or Deformable by abandoning background tokens with weak semantics. However, these methods are suboptimal in judging background information and lack enhanced attention to more fine-grained features.

\section{Methodology}
We first describe the overall architecture of Focus-DETR. Then, we elaborate on our core contributions: (a) Constructing a scoring mechanism that considers both localization and category semantic information from multi-scale features. Thus we obtain two-level explicit discrimination for foreground and fine-grained object tokens; (b) Based on the scoring mechanism, we feed tokens with different semantic levels into the encoder with dual attention, which enhances the semantic information of queries and balances model performance and calculation cost. A detailed analysis of the computational complexity is provided.

\subsection{Model Architecture}
As shown in Fig.~\ref{fig:overview}, Focus-DETR is composed of a backbone, a encoder with dual attention and a decoder. The backbone can be equipped with ResNet~\cite{ResNet} or Swin Transformer~\cite{SwinTransformer}. To leverage multi-scale features $\{\boldsymbol{f_l}\}_{l=1}^{L}~(L=4)$ from the backbone, where $\boldsymbol{f_l} \in \mathbb{R}^{C \times H_l \times W_l}$, we obtain the feature maps $\{ \boldsymbol{f_1,f_2,f_3}\}$ in three different scales~($i.e.$, 1/8, 1/16, 1/32) and downsample $\boldsymbol{f_3}$ to get $\boldsymbol{f_4}$ ($i.e.$, 1/64).

Before being fed into the encoder with dual attention, the multi-scale feature maps $\{\boldsymbol{f_l}\}_{l=1}^{L}$ first go through a foreground token selector (Section \ref{fts}) using a series of top-down score modulations to indicate whether a token belongs to the foreground. Then, the selected foreground tokens of each layer will pass through a multi-category score predictor to select tokens with higher objectiveness score by leveraging foreground and semantic information~(Section \ref{fts}). These object tokens will interact further with each other and complement the semantic limitation of the foreground queries through the proposed  dual attention (Section \ref{oesm}).


\subsection{Scoring mechanism} \label{fts}
\label{sec:sfp}
\begin{figure}[t]
	\begin{center}
		\includegraphics[width=1.0\linewidth]{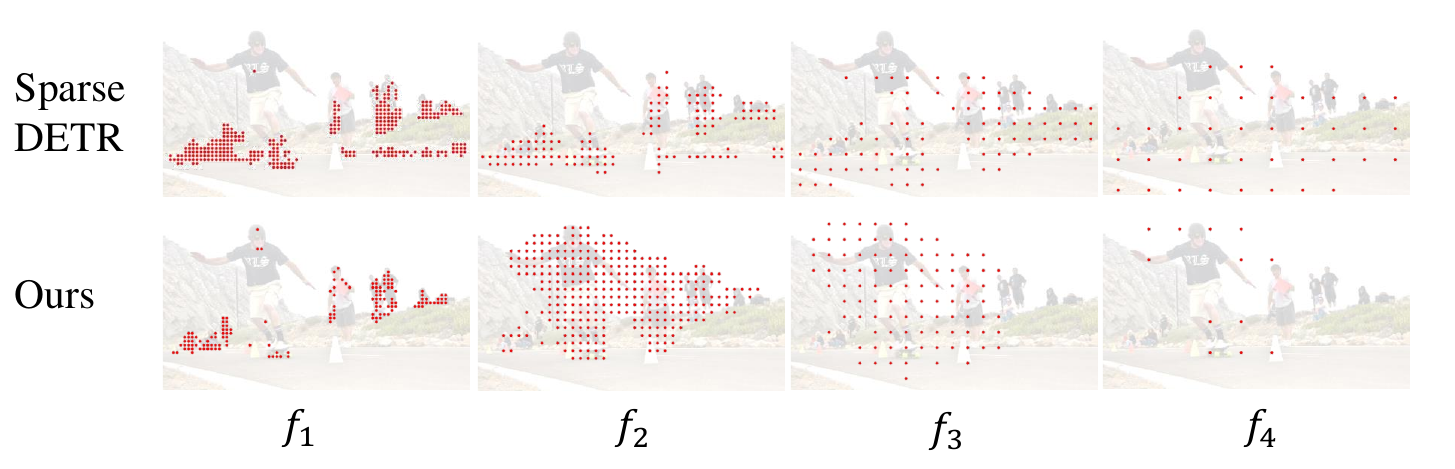}
	\end{center}
	\vspace{-0.7em}
	\caption{The foreground tokens preserved in different feature maps of Sparse DETR and our Focus-DETR. The red dots indicate the position of the reserved token corresponding to the original image based on the stride.}
	\label{fig:vision3}

\end{figure}

\textbf{Foreground Token Selector.} Sparse DETR\cite{sparse_DETR} has demonstrated that only involving a subset of tokens for encoders can achieve comparable performance. However, as illustrated in Fig.~\ref{fig:vision3}, the token selection provided by Sparse DETR~\cite{sparse_DETR} has many drawbacks. In particular, many preserved tokens do not align with foreground objects.

We think the challenge from Sparse DETR lies in that its supervision of token selection relies on DAM. The correlation between DAM and retained foreground tokens will be reduced due to learnable queries, which brings errors during training. Instead of predicting pseudo-ground truth~\cite{sparse_DETR}, we leverage ground truth boxes and labels to supervise the foreground selection inspired by~\cite{FocalLoss}. To properly provide a binary label for each token on whether it appears in foreground, we design a label assignment protocol to leverage the multi-scale features for objects with different scales. 

In particular, we first set a range of sizes for the bounding boxes of different feature maps, and add the overlap of the adjacent interval by 50\% to enhance the prediction near boundary. Formally, for each token $t_l^{(i,j)}$ with stride $s_l$, where $l$ is the index of scale level, and $(i, j)$ is the position in the feature map, we denote the corresponding coordinate $(x,y)$ in the original image as $\left(\left\lfloor\frac{s_l}{2}\right\rfloor + i \cdot s_l,\left\lfloor\frac{s_l}{2}\right\rfloor + j \cdot s_l\right)$. Considering the adjacent feature map, our protocol determines the label $l_l^{(i,j)}$ according to the following rules, \textit{i.e.,}
\begin{equation}
	\label{eq6}
	l_l^{(i,j)}=\left\{
	\begin{aligned}
		1 & , & (x,y) \in \mathcal{D}_{Bbox} ~\wedge~d_l^{(i,j)}\in[r^l_b,r_e^l] \\
		0 & , & (x,y) \notin \mathcal{D}_{Bbox} ~\vee~d_l^{(i,j)}\notin[r^l_b,r_e^l]
	\end{aligned},
	\right.
\end{equation}

\noindent where $\mathcal{D}_{Bbox}~(x,y,w,h)$ denotes the ground truth boxes, $d_l^{(i,j)}$=$max(\frac{h}{2},\frac{w}{2})\in[r_b^l,r_e^l]$, represents the maximum checkerboard distance between $(x,y)$ and the bounding box center,  $[r^l_b,r_e^l]$ represents the interval of object predicted by the $l$-layer features and $r_b^l<r_b^{l+1}<r_e^l<r_e^{l+1}$ and $r_b^{l+1}=\frac{(r_b^l+r_e^l)}{2}$, $l=\{0,1,2,3\}$, $r_b^0=0$ and $r_e^3=\infty$.

\begin{figure}[t]
	\begin{center}
		\includegraphics[width=0.9\linewidth]{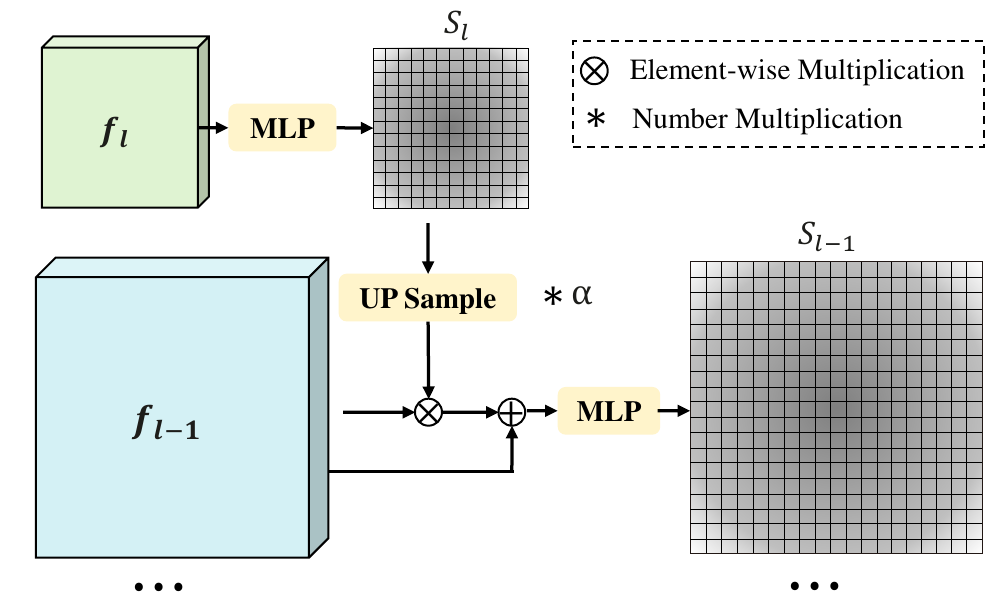}
	\end{center}
	\vspace{-1.2em}
	\caption{The operation of top-down score modulation. For multi-scale feature maps, we use a shared MLP to calculate $\{S_1, S_{2}, ...\}$. $S_l$ is incorporated  in the calculation of $S_{l-1}$ by a dynamic coefficient $\alpha$ and feature map $\boldsymbol{f_{l-1}}$.}
	\label{fig:constrain}
	\vspace{-1.1em}
\end{figure}

Another drawback of DETR sparse methods is the insufficient utilization of multi-scale features. In particular, the semantic association and the discrepancy in the token selection decisions between different scales are ignored. To fulfill this gap, we construct the FTS module with top-down score modulations. We first design a score module based on Multi-Layer Perceptron~(MLP) to predict the foreground score in each feature map. Considering that high-level feature maps contain richer semantic than low-level features with higher resolution, we leverage the foreground score of high-level semantics as complement information to modulate the feature maps of adjacent low-level semantics. As shown in Fig.~\ref{fig:constrain}, our top-down score modulations only transmits foreground scores layer by layer through upsampling. Formally, given the feature map $\boldsymbol{f_l}$ where $l \in \{2, 3, 4\}$,
\begin{equation}
	\begin{aligned}
		& S_{l-1} = \mathbf{MLP_F}(\boldsymbol{f_{l-1}}(1 + \mathbf{UP}(\alpha_l * S_l))),
	\end{aligned}
\end{equation}
where $S_l$ indicates the foreground score of the $l$-th feature map, $\mathbf{UP(\cdot)}$ is the upsampling function using bilinear interpolation, $\mathbf{MLP_F(\cdot)}$ is a global score predictor for tokens in all the feature maps, $\{\alpha_l\}_{l=1}^{L-1}$ is a set of learnable modulation coefficients, and $L$ indicates the layers of multi-scale feature maps.
The localization information of different feature maps is correlated with each other in this way.

\textbf{Multi-category score predictor}. After selecting tokens with a high probability of falling in the foreground, we then seek an efficient operation to determine more fine-grained tokens for query enhancement with minimal computational cost. Intuitively, introducing more fine-grained category information would be beneficial in this scenario. Following this motivation, we propose a novel more fine-grained token selection mechanism coupled with the foreground token selection to make better use of the token features. As shown in Fig.~\ref{fig:overview}, to avoid meaningless computation of the background token, we employ a stacking strategy that considers both localization information and category semantic information. Specifically, the product of foreground score and category score calculated by a predictor $\mathbf{MLP_C(\cdot)}$ will be used as our final criteria $p_j$ for determining the fine-grained tokens involved in the attention calculation, \textit{i.e.}, 
\begin{equation}
	\begin{aligned}
		& p_j = s_j\times c_j = s_j\times \mathbf{MLP_C}(T_f^j),
	\end{aligned}
\end{equation}
where $s_j$ and $c_j$ represent foreground score and category probabilities of $T_f^j$ respectively. Unlike the query selection strategy of two-stage Deformable DETR~\cite{Deformable_DETR} from the encoder's output, our multi-category probabilities do not include background categories~($\emptyset$). We will determine the tokens for enhanced calculation based on the $p_j$.

\subsection{Calculation Process of Dual Attention}   \label{oesm}
The proposed reliable token scoring mechanism will enable us to perform more fine-grained and discriminatory calculations. After the foreground and fine-grained object tokens are gradually selected based on the scoring mechanism, we first exploit the interaction information of the fine-grained object tokens and corresponding position encoding by enhanced self-attention. Then, the enhanced object tokens will be scattered back to the original foreground tokens. This way, Focus-DETR can leverage the foreground queries with enhanced semantic information. In addition, because of reliable fine-grained token scoring, dual attention in Encoder effectively boosts the performance with only a negligible increase in calculation cost compared to the unsophisticated query sparse strategy. We utilize Algorithm~\ref{alg:encoder} to illustrate the fine-grained feature selection and enhancement process in the encoder with dual attention. 
\begin{algorithm}[t]
	\caption{Encoder with Dual Attention} 
	\label{alg:encoder}
	\begin{algorithmic}[1]
		\REQUIRE All tokens $T_a$,~foreground tokens $T_f$,~position embedding $PE_f$,~object token number $k$,~foregroud score $~S_f$, foreground token index $I_f$
		\ENSURE all tokens $T_a'$ and foreground tokens $T_f'$ after one encoder layer
		\STATE category score $C_f~\leftarrow~\mathbf{MLP_C}(T_f)$
		\STATE maximum of category score $S_c~\leftarrow~max(C_f)$
		\STATE object token score $S_p~=~S_c \cdot S_f$
		\STATE $Idx_k^{obj}~\leftarrow~\mathbf{TopK}(S_p, k)$
		\STATE $~T_o~\leftarrow~T_f[Idx_k^{obj}],~PE_o~\leftarrow~PE_f[Idx_k^{obj}]$
		\STATE $q=k=PE_o+T_o,~v=T_o$
		\STATE $T_o\leftarrow\mathbf{MHSA}(q,~k,~v)$
		\STATE $T_o\leftarrow\mathbf{Norm}(v+T_o)$
		\STATE update $T_o$ in $T_f$ according to $Idx_k^{obj}$
		\STATE $q~=~T_f',~k~=~T_a + PE_f,~v~=~T_a$
		\STATE $T_f'~\leftarrow~\mathbf{MSDeformAttn}(q,k,v)$
		\STATE update $T_f'$ in $T_a$ according to $I_f$
	\end{algorithmic}
\end{algorithm}
\subsection{Complexity Analysis} 
We further analyze the results in Fig.~\ref{fig:flops} and our claim that the fine-grained tokens enhanced calculation adds only a negligible calculation cost mathematically. We denote the computational complexity of deformable attention in the encoder and decoder as \{$G_{DA}^e$, $G_{DA}^d$\}, respectively. We calculate $G_{DA}$ with reference to Deformable DETR~\cite{Deformable_DETR} as follows: 
\begin{equation}
	\begin{aligned}
		& G_{DA} = O(KC+3MK+C + 5K)N_qC , \\
	\end{aligned}
\end{equation}
where $N_q$~($N_q\leq HW=\sum_{i=1}^{L}h_iw_i$) is the number of queries in encoder or decoder, $K$ is the sampling number and $C$ is the embedding dims. For encoder, we set $N_{qe}$ as $\gamma HW$, where $\gamma$ is the ratio of preserved foreground tokens. For decoder, we set $N_{qd}$ to be a constant. In addition, the complexity of the self-attention module in decoder is $O(2N_{qd}C^2+N_{qd}^2C)$.~For an image whose token number is approximately $1\times 10^4$, $\frac{G_{DA}^e}{G_{DA}^d}$ is approximately $7$ under the common setting $\{K=4, C=256, N_{qd}=900, \gamma=1\}$. When $\gamma$ equals  0.3, the calculation cost in the Transformer part will reduce over 60\%. This intuitive comparison demonstrates that the encoder is primarily responsible for redundant computing. Then we define the calculation cost of the fine-grained tokens enhanced calculation as $G_{OEC}$:
\begin{equation}
	\begin{aligned}
		& G_{OEC} = O(2N_0C^2+N_0^2C) ,\\
	\end{aligned}
\end{equation}
where $N_0$ represents the number of fine-grained tokens that obtained through scoring mechanism. When $N_0=300$, $\frac{G_{OEC}}{(G_{DA}^e+ G_{DA}^d)}$ is only less than 0.025, which has a negligible impact on the overall model calculation. 

\subsection{Optimization}
Like DETR-like detectors, our model is trained in an end-to-end manner,
and the loss function is defined as:
\begin{equation}
	\begin{aligned}
		\mathcal{L} =\lambda_m\widehat{\mathcal{L}}_{match}+\lambda_d\widehat{\mathcal{L}}_{dn}+\lambda_f\widehat{\mathcal{L}}_{f}+\lambda_e\widehat{\mathcal{L}}_{enc}~,
	\end{aligned}
\end{equation}
where $\widehat{\mathcal{L}}_{match}$ is the loss for pair-wise matching based on Hungarian algorithm, $\widehat{\mathcal{L}}_{dn}$ is the loss for denoising models, $\widehat{\mathcal{L}}_{f}$ is the loss for foreground token selector, $\widehat{\mathcal{L}}_{enc}$ is the loss for auxiliary optimization through the output of the last encoder layer, $\lambda_m$, $\lambda_d$, $\lambda_f$, $\lambda_a$ are scaling factors.
\begin{table*}[htbp]
	\small
	\centering
	\setlength{\tabcolsep}{2.5mm}
	\begin{tabular}{c|c|cccccc|ccc}
		\toprule
		Model & Epochs & AP& $AP_{50}$& $AP_{75}$& $AP_{S}$& $AP_{M}$& $AP_{L}$     & Params& GFLOPs & FPS \\
		\midrule
		Faster-RCNN\cite{Faster_RCNN} & 108& 42.0 & 62.4& 44.2 & 20.5& 45.8 & 61.1& 42M & 180&25.3\\
		DETR(DC5)\cite{DETR} &  500& 43.3 & 63.1& 45.9 & 22.5& 47.3 & 61.1& 41M & 187&11.2\\
		
		Efficient-DETR\cite{efficient_DETR} &  36& 44.2 &62.2& 48.0& 28.4& 47.5 &56.6&32M&159&--\\
		
		Anchor-DETR-DC5\cite{anchor_DETR} &  500& 44.2& 64.7 &47.5 &24.7 &48.2 &60.6&-- &--&19.0\\
		PnP-DETR($\alpha=0.33$)\cite{pnp_DETR} & 500& 42.7& 62.8 &45.1 &22.4& 46.2& 60& -- & --&42.5\\
		Conditional-DETR-DC5\cite{conditional_DETR}  & 108& 45.1& 65.4& 48.5& 25.3& 49.0& 62.2& 44M & 195&11.5\\
		Conditional-DETR-V2\cite{conditional_DETR_v2}  & 50& 44.8& 65.3& 48.2& 25.5& 48.6& 62.0& 46M & 161&--\\

		Dynamic DETR(5 scales)\cite{DynamicDETR} &  50& 47.2& 65.9& 51.1& 28.6 &49.3 &59.1& 58M & --&--\\
		DAB-Deformable-DETR\cite{DAB_DETR} &  50&46.9 &66.0& 50.8& 30.1& 50.4 &62.5& 44M & 256&14.8\\
		
		UP-DETR\cite{upDETR} &  300& 42.8& 63.0& 45.3& 20.8& 47.1 &61.7& -- & --&--\\
		SAM-DETR\cite{SAMDETR} &  50 &45.0& 65.4& 47.9& 26.2& 49.0 &63.3& 58M & 210&24.4 \\

		Deformable DETR\cite{Deformable_DETR} &  50 & 46.2& 65.2& 50.0& 28.8& 49.2& 61.7&40M&173&19.0\\
		Sparse DETR($\alpha=0.3$)\cite{sparse_DETR} &  50&46.0& 65.9& 49.7& 29.1& 49.1& 60.6& 41M & 121&23.2\\

		DN-Deformable-DETR\cite{DN_DETR} &  50& 48.6& 67.4& 52.7 &31.0& 52.0& 63.7& 48M & 265&18.5\\

		\midrule
		DINO\cite{DINO}&36&50.9&69.0&55.3&34.6&54.1&64.6&47M&279&14.2\\
		 + Sparse DETR($\alpha=0.3$) &{36}&48.2&{65.9}&{52.5}&{30.4}&{51.4}&{63.1}&{47M}&{152}&{20.2}		
		\\
		or + \textbf{Focus-DETR (Ours)($\alpha=0.3$)} &{36}&\bf{50.4}&{68.5}&{55.0}&{34.0}&{53.5}&{64.4}&{48M}&{154}&{20.0}		
		\\
		
		\bottomrule
	\end{tabular}%
	\vspace{-0.5em}
	\caption{Results for our Focus-DETR and other detection models with the ResNet50 backbone on COCO val2017. Herein, $\alpha$ indicates the $keep~ratio$ for methods that prune background tokens. All reported FPS are measured on a NVIDIA V100. }
	\label{tab:compare}%
	\vspace{-1.5em}
\end{table*}%

\textbf{Loss for feature scoring mechanism.}
Focus-DETR obtains foreground tokens by the FTS module. Focal Loss~\cite{FocalLoss} is applied to train FTS as follow:
\begin{equation}
	\begin{aligned}
		\widehat{\mathcal{L}}_{f}=-\alpha_f(1-p_f)^\gamma log(p_f)~,
	\end{aligned}
\end{equation}
where $p_f$ represents foreground probability, $\alpha_f = 0.25$ and $\gamma = 2$ are empirical hyperparameters.


\section{Experiments}
\subsection{Experimental Setup}
\textbf{Dataset}: We conduct experiments on the
challenging COCO 2017~\cite{COCO} detection dataset, which contains 117K training images and 5K validation images. Following
the common practice, we report the standard average
precision (AP) result on the COCO validation dataset.

\textbf{Implementation Details}: The implementation details of
Focus-DETR mostly align with the original model in detrex~\cite{ren2023detrex}. We
adopt ResNet-50~\cite{ResNet}, which is pretrained using ImageNet~\cite{ImageNet} as the backbone and train our model with $8\times$Nvidia V100 GPUs using the AdamW~\cite{Adam} optimizer. In addition, we perform experiments with ResNet-101 and Swin Transformer as the backbone. The initial learning rate is set as $1\times 10^{-5}$ for the backbone and $1\times 10^{-4}$ for the Transformer encoder-decoder framework, along with a weight decay of $1\times 10^{-4}$. The learning rate decreases at a later stage by 0.1. The batch size per GPU is set to 2. For the scoring mechanism, the loss weight coefficient of the FTS is set to 1.5. The  $\mathbf{MLP_C(\cdot)}$ shares parameters with the corresponding in the decoder layer and is optimized along with the training of the entire network.
In addition, we decrease the cascade ratio by an approximate arithmetic sequence, and the lower threshold is 0.1. We provide more detailed hyper-parameter settings in Appendix~\ref{sec:cascade}, including the reserved token ratio in the cascade structure layer by layer and the object scale interval for each layer.
\begin{table}[htbp]
	\scriptsize
	\centering
	\setlength{\tabcolsep}{1.0mm}
	
	\begin{tabular}{c|c|ccc|cc}
		\toprule
		Model  & Epochs  & $AP$    & $AP_{50}$&$AP_{75}$   & Params & GFLOPs \\
		\midrule
		
		
		Faster RCNN-FPN~\cite{Faster_RCNN}&108&44.0&63.9&47.8&60M&246\\
		DETR-DC5~\cite{DETR}&500&44.9&64.7& 47.7&60M&253\\
		Anchor-DETR*~\cite{anchor_DETR}&50&45.1&65.7&48.8&58M&--\\
		DN DETR~\cite{DN_DETR}&50&45.2&65.5& 48.3&63M&174\\
		DN DETR-DC5~\cite{DN_DETR}&50&47.3& 67.5& 50.8& 63M&282\\
		Conditional DETR-DC5~\cite{conditional_DETR}&108&45.9& 66.8& 49.5&63M&262\\
		DAB DETR-DC5~\cite{DAB_DETR}&50&46.6& 67.0& 50.2&63M&296\\
		\midrule
		{\textbf{Focus-DETR~(Ours)}}&36& {51.4}& { 70.0}& {55.7}& {67M}& {221}\\
		\bottomrule
	\end{tabular}%
	\vspace{-1.2em}
	\caption{Comparison of Focus-DETR~(DINO version) and other models with ResNet101 backbone. Our Focus-DETR preserve 30\% tokens after the backbone. The models with superscript * use 3 pattern embeddings.}
	\vspace{-1.4em}
	\label{tab:focus-backbone}%
\end{table}
\begin{table}[htbp]
	\scriptsize
	\centering
	\setlength{\tabcolsep}{1.5mm}
	\begin{tabular}{c|c|c|c|c}
		\toprule
		Model&AP& Corr&GFLOPs&FPS  \\
		\hline
		Deformable DETR~\textcolor{blue}{(priori)}&46.2 &--&177&19 \\
		+ Sparse DETR~($\alpha=0.3$)& 46.0&0.7211$\pm$0.0695&121&23.2 \\
		or +\textbf{ Focus-DETR}~($\alpha=0.3$)&46.6 &--&123& 23.0\\
		\hline
		Deformable DETR~\textcolor{blue}{(learnable)}&45.4 &--&173&19 \\
		+ Sparse DETR~($\alpha=0.3$)&43.5 &0.5081$\pm$0.0472&118&24.2 \\
		or + \textbf{Focus-DETR}~($\alpha=0.3$)&45.2 &--&120&23.9 \\
		\hline
		DN-Deformable-DETR~\textcolor{blue}{(learnable)}&48.6 &--&195&18.5 \\
		+ Sparse DETR~($\alpha=0.3$)&47.4 &0.5176$\pm$0.0452&137&23.9 \\
		or + \textbf{Focus-DETR}~($\alpha=0.3$)&48.5 &--&138&23.6 \\
		\hline
		DINO~\textcolor{blue}{(mixed)}& 50.9&--&279&14.2 \\
		
		+ Sparse DETR~($\alpha=0.3$)&48.2 &0.5784$\pm$0.0682&152&20.2 \\
		or + \textbf{Focus-DETR}~($\alpha=0.3$)&50.4 &--&154&20.0 \\
		\bottomrule
	\end{tabular}%
	\vspace{-1.2em}

	\caption{ \textbf{$Corr$}: the correlation of DAM and retained foreground(5k validation set). \textbf{``priori''}: position and content query~(encoder selection); \textbf{``learnable''}: position and content query~(initialization); \textbf{``mixed''}: position query~(encoder selection), content query~(initialization).}
	
	\small \label{tab:correlation}%
	\vspace{-1.7em}
\end{table}

\subsection{Main Results}
Benefiting from the well-designed scoring mechanisms towards the foreground and more fine-grained object tokens, Focus-DETR can focus attention on more fine-grained features, which further improves the performance of the DETR-like model while reducing redundant computations. 

Table~\ref{tab:compare} presents a thorough comparison of the proposed Focus-DETR~(DINO version) and other DETR-like detectors~\cite{DETR, efficient_DETR, Deformable_DETR, anchor_DETR, pnp_DETR, conditional_DETR, conditional_DETR_v2, SMCA-R, TSP-RCNN-R, DynamicDETR, DAB_DETR, DN_DETR, upDETR, SAMDETR, sparse_DETR}, as well as Faster R-CNN~\cite{Faster_RCNN}. We compare our model with efficient DETR-based detectors~\cite{pnp_DETR, sparse_DETR}, our Focus-DETR with keep-ratio of 0.3 outperforms PnP-DETR~\cite{pnp_DETR}~(+7.9 AP). We apply the Sparse DETR to DINO to build a solid baseline. Focus-DETR outperforms Sparse DETR~(+2.2 AP) when embedded into DINO. When applied to the DINO~\cite{DINO} and compared to original DINO, we lose only 0.5 AP, but the computational cost is reduced by 45\% and the inference speed is improved 40.8\%.

\begin{figure*}
	\begin{center}
		\includegraphics[width=0.87\textwidth]{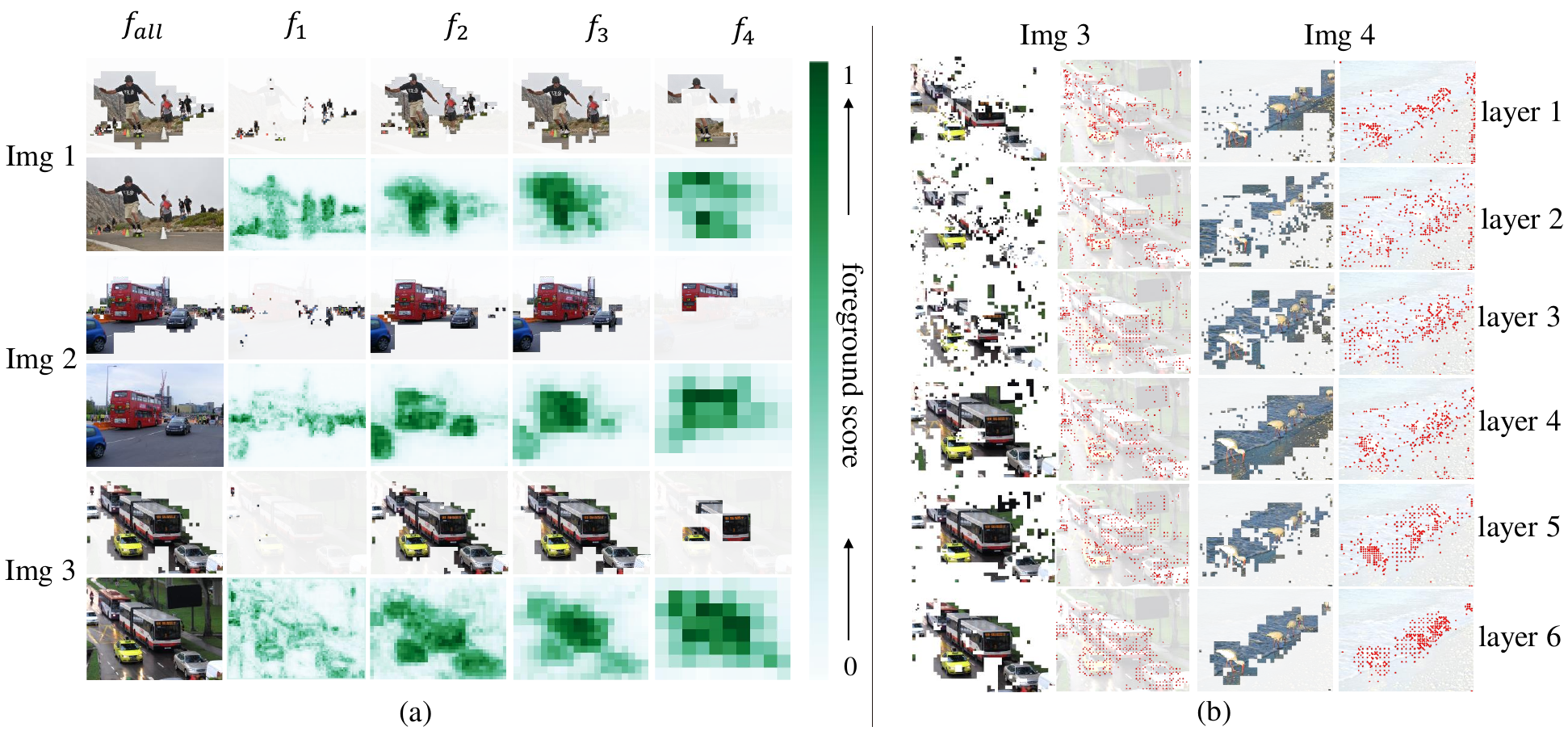}
	\end{center}
	\vspace{-1.3em}
	\caption{ Visualization results of preserved foreground tokens distribution at multi-scale feature maps as shown (a) and $k$ object tokens evolution at different encoder layers as shown (b). $\{$Img1, Img2, Img3, Img4$\}$ represent four test images, $\{\boldsymbol{f_1}$, $\boldsymbol{f_2}$, $\boldsymbol{f_3}$, $\boldsymbol{f_4}\}$ represent foreground tokens at four feature maps, $\{$layer~1, layer~2 ...$\}$ represent different encoder layers.}
	\label{fig:vision2}
	\vspace{-1.7em}
\end{figure*}

\begin{figure}[t]
	\begin{center}
		\includegraphics[width=0.8\linewidth]{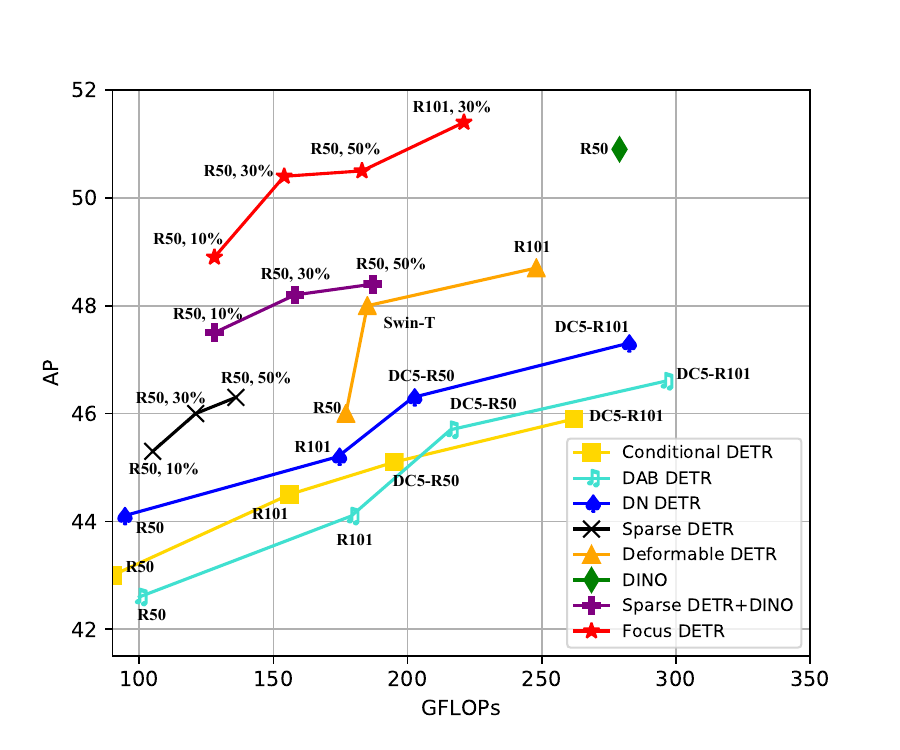}
	\end{center}
	\vspace{-1.2em}
	\caption{ Performance of recent object detectors in terms of average precision(AP) and GFLOPs. The GFLOPs is measured using 100 validation images.}
	\label{fig:flopsap}
	\vspace{-1.6em}
\end{figure}
In Fig.~\ref{fig:flopsap}, we plot the AP with GFLOPs to provide a clear picture of the trade-off between accuracy and computation cost. Overall, Our Focus-DETR~(DINO version) achieve state-of-the-art performance when compared with other DETR-like detectors. 

To verify the adaptability of Focus-DETR to the stronger backbone ResNet-101~\cite{ResNet} and the effect of the ratio of the preserved foreground on model performance, we perform a series of extensive experiments. As shown in Table~\ref{tab:focus-backbone}, compared to other DETR-like models~\cite{DAB_DETR, DN_DETR, anchor_DETR, DETR, SMCA-R, TSP-RCNN-R, Faster_RCNN}, Focus-DETR (DINO version) achieves higher AP with fewer GFLOPs. Moreover, using a Swin Transformer pretrained on ImageNet as backbone, we also achieve excellent performance, as shown in Appendix~\ref{swin_backbone_ana}.

\subsection{Extensive Comparison}
Sparse DETR is state-of-the-art for lightweight DETR-like models.  As mentioned earlier, sparse DETR will cause significant performance degradation when using learnable queries. To verify the universality of Focus-DETR, we compare our model with excellent and representative DETR-like models equipped with Sparse DETR, including Deformable DETR~\cite{Deformable_DETR}, DN-DETR~\cite{DN_DETR} and DINO~\cite{DINO}. 

In addition to the Sparse DETR, we apply the Sparse DETR to Deformable DETR(two-stage off), DN-Deformable DETR and DINO to construct three baselines. We retain all the Sparse DETR's designs for a  fair enough comparison, including the auxiliary encoder loss and related loss weight. We also optimize these baselines by adjusting hyperparameters to achieve the best performance. As shown in Table \ref{tab:correlation}, when applying Sparse DETR to Deformable DETR without two-stage, DN-Deformable-DETR and DINO, the AP decreases 1.9, 1.2 and 2.7. We calculate $Corr$ proposed by Sparse DETR that denotes the correlation bewteen DAM and selected foreground token, we calculate the top 10\% tokens to compare the gap more intuitively. As shown in Table~\ref{tab:correlation}, their $Corr$s are far lower than original Sparse DETR, which means foreground selector does not effectively learn DAM. Compared to Sparse DETR, Focus-DETR achieves 1.7, 1.1 and 2.2 higher AP with similar latency in Deformable DETR(two-stage off), DN-Deformable DETR and DINO.

As shown in Fig.~\ref{fig:overview}, it seems that our encoder using dual attention can be independently embedded into Sparse DETR or other DETR-like models. However, a precise scoring mechanism is critical to dual attention. We added the experiments of applying the encoder with dual attention to Sparse DETR in Appendix~\ref{sec:dual_attn_model}. Results show us that fine-grained tokens do not bring significant performance gains.

\subsection{Ablation Studies}
We conduct ablation studies to validate the effectiveness of our proposed components. Experiments are performed with ResNet-50 as the backbone using 36 epochs.



\textbf{Effect of foreground token selection strategy.}
Firstly, simply obtaining the token score using a foreground score predictor without supervision  achieves only 47.8 AP and is lower than that~(48.2 AP) of DINO pruned by Sparse DETR. As shown in the second row of Table~\ref{tab:ablation1}, by adding supervision with our improved label assignment strategy, Focus-DETR yields a significant improvement of +1.0 AP. In addition, top-down score modulations optimize the performance of FTS  by enhancing the scoring interaction between multi-scale feature maps. As shown in the third row of Table~\ref{tab:ablation1}, Focus-DETR equipped with the top-down score modulation achieves +0.4 AP. ~As the visualization shown in Fig.~\ref{fig:vision2}~(a), we can observe that our method precisely select the foreground tokens. Moreover, feature maps in different levels tend to focus on objects with different scales. Furthermore, we find that that there is an overlap between the object scales predicted by adjacent feature maps due to our scale overlap setting. We provide more detailed overlap setting details in the Appendix~\ref{sec:overlap}.

\begin{table}[htbp]
	\scriptsize
	\centering
	\setlength{\tabcolsep}{0.7mm}
	\vspace{-1.2em}
	\begin{tabular}{cc|c|c|c|c|c|c|c}
		\toprule
		\multicolumn{2}{c|}{FTS}                     & \multicolumn{1}{c|}{\multirow{1}{*}{score}} & \multicolumn{1}{c|}{\multirow{2}{*}{cascade}} & \multicolumn{1}{c|}{\multirow{1}{*}{dual}} & \multicolumn{1}{c|}{\multirow{2}{*}{AP}} & \multicolumn{1}{c|}{\multirow{2}{*}{$AP_{50}$}} & \multicolumn{1}{c|}{\multirow{2}{*}{$AP_{75}$}} &   \multicolumn{1}{c}{\multirow{2}{*}{FPS}} \\ \cline{1-2}
		\multicolumn{1}{c|}{predictor} & supervision & \multicolumn{1}{c|}{modulations}                                 & \multicolumn{1}{c|}{}                         & \multicolumn{1}{c|}{attention}                      & \multicolumn{1}{c|}{}                    & \multicolumn{1}{c|}{}                      & \multicolumn{1}{c|}{}                      & \multicolumn{1}{c}{}                                             \\ \midrule
		\multicolumn{1}{c}{$\checkmark$}  &  &   &    &  &47.8& 65.2 & 52.1 &20.4   \\ 
		\multicolumn{1}{c}{$\checkmark$}         & $\checkmark$ &   & &&48.8   &66.2   & 53.2                      &20.4  \\ 
		\multicolumn{1}{c}{$\checkmark$}         & $\checkmark$           & $\checkmark$   &     & & 49.2 & 66.4 &53.7 &20.3\\ 
		\multicolumn{1}{c}{$\checkmark$} & $\checkmark$           & $\checkmark$& $\checkmark$  &  &49.7  &66.9  &54.1 &20.3 \\ 
		\multicolumn{1}{c}{$\checkmark$}& $\checkmark$  & $\checkmark$& $\checkmark$ & $\checkmark$& \textbf{50.4}& \textbf{68.5} & \textbf{55.0 } & 20.0\\ \bottomrule
	\end{tabular}
	\vspace{-1.0em}
	\caption{Ablation studies on the FTS and dual attention. FTS is the foreground token selector. Dual attention represents the our encoder structure. Supervision indicates the label assignment from the ground truth boxes.}
	\label{tab:ablation1}
	\vspace{-1.2em}
\end{table}

\textbf{Effect of cascade token selection.} When keeping a fixed number of tokens in the encoder,  the accumulation of pre-selection errors layer by layer is detrimental to the detection performance. To increase the fault tolerance of the scoring mechanism,  we design the cascade structure for the encoder to reduce the number of foreground tokens layer by layer~(Section \ref{sec:sfp}). As shown in Fig.~\ref{fig:vision2} (b), we can see the fine-grained tokens focusing process in the encoder as the selecting range decreases, which enhances the model's fault tolerance and further improves the model's performance. As illustrated in the fourth row of Table~\ref{tab:ablation1}, Focus-DETR equipped with cascade structure achieves +0.5 AP.

\textbf{Effect of the dual attention.} Unlike only abandoning the background tokens, one of our contributions is reconstructing the encoder using dual attention  with negligible computational cost. Tokens obtained after the enhanced calculation supplement the semantic weakness of the foreground queries due to the limitation in distant token mixing. We further analyze the effects of the encoder with dual attention. As shown in the fifth row of Table~\ref{tab:ablation1},~the encoder with dual attention brings +0.8 AP improvement.
These results demonstrate that enhancing fine-grained tokens is beneficial to boost detection performance and the effectiveness of our stacked position and semantic information for fine-grained feature selection, as shown in Fig.~\ref{fig:patchvisual}.
	
\begin{table}[htbp]
	\scriptsize
	\centering
	\setlength{\tabcolsep}{1.0mm}
	\normalsize
	\scriptsize
	\centering
	\setlength{\tabcolsep}{4mm}
	\vspace{-1.2em}
	\begin{tabular}{cc|c|c|c}
		\toprule
		\multicolumn{1}{c|}{Top-down} & Bottom-up & \multicolumn{1}{c|}{$AP$}                    & \multicolumn{1}{c|}{$AP_{50}$}                      & \multicolumn{1}{c}{$AP_{75}$}                                      \\ \midrule
		
		\multicolumn{1}{c}{}         &           &  49.7 & 66.9 &54.0 \\ 
		\multicolumn{1}{c}{$\checkmark$}        &  &\textbf{50.4}& \textbf{68.5} &\textbf{ 55.0} \\ 
		\multicolumn{1}{c}{}         & $\checkmark$         & 50.2& 68.4 & 54.6\\ \bottomrule
	\end{tabular}
	\vspace{-1.0em}
	\caption{Association methods between scores of multi-scale feature maps. We try top-down and bottom-up modulations.}
	\label{tab:ablation2}
	\vspace{-1.7em}
\end{table}

\textbf{Effect of top-down score modulation.}
We further analysis the effect of the multi-scale scoring guidance mechanisms in our method.As shown in Table~\ref{tab:ablation2}, we can observe that utilizing multi-scale information for score prediction brings consistent improvement (+0.5 or +0.7 AP). We also conduct ablation experiments for different score modulation methods. The proposed top-down score guidance strategy~(Section~\ref{sec:sfp}) achieves 0.2 higher AP than bottom-down strategy, which justifies our motivation that using high-level scores to modulating low-level foreground probabilities is beneficial for the final performance. 



\textbf{Effect of pruning ratio.}
As shown in Table~\ref{tab:keep-ratio}, we analyze the detection performance and model complexity when changing the ratio of foreground tokens retained by different methods. 
Focus-DETR achieves optimal performance when keeping the same ratio. Specifically, Focus-DETR achieves +2.7 AP than Sparse DETR and +1.4AP than DINO equipped with Sparse DETR's strategies with similar computation cost at 128 GFLOPs.

\begin{table}[htbp]
	\scriptsize
	\centering
	\setlength{\tabcolsep}{1.5mm}
	\vspace{-1.2em}
	\begin{tabular}{c|c|cccc|c|c}
		\toprule
		Model  & $\alpha$  & $AP$      & $AP_S$ & $AP_M$ & $AP_L$ & GFLOPs&FPS \\
		\hline
		\multirow{3}[2]{*}{Sparse DETR~\cite{sparse_DETR}}
		&0.1&45.3&  28.4& 48.3& 60.1&105&25.4\\
		&0.2&45.6&  28.5& 48.6& 60.4&113&24.8\\
		&0.3&46.0&  29.1& 49.1& 60.6&121&23.2\\
		\multirow{1}[2]{*}{(epoch=50)}
		&0.4&46.2& 28.7& 49.0& 61.4&128&21.8\\
		&0.5&46.3& 29.0& 49.5& 60.8&136&20.5\\
		\hline
		\multirow{2}[2]{*}{DINO~\cite{DINO} }
		&0.1&47.5& 29.1&50.7&62.7&126&23.9\\
		&0.2&47.9& 30.0&51.1&62.9&139&21.4\\
		+ Sparse DETR~\cite{sparse_DETR}	&0.3&48.2&30.5&51.4&63.1&152&20.2\\
		\multirow{1}[2]{*}{(epoch=36)}
		&0.4&48.4& 30.5&51.8&63.2&166&18.6\\
		&0.5&48.4& 30.6&51.8&63.4&181&18.1\\
		\hline
		\multirow{3}[2]{*}{\textbf{Focus-DETR}}
		&0.1&48.9&32.6&52.6&64.1&128&23.7\\
		&0.2&49.8&32.3&52.9&64.0&141&21.3\\
		&0.3&50.4&33.9&53.5&64.4&154&20.0\\
		\multirow{1}[2]{*}{(epoch=36)}
		&0.4&50.4&34.0&53.7&64.1&169&18.5\\
		&0.5&50.5&34.4&53.8&64.0&183&17.9\\
		\bottomrule
	\end{tabular}%
\vspace{-1.3em}
\caption{Experiment results in performance and calculation cost when changing the ratio of foreground tokens retained by Focus-DETR, Sparse DETR, and DINO+Sparse DETR.}
	
	\label{tab:keep-ratio}%
	\vspace{-1.9em}
\end{table}

\subsection{Limitation and Future Directions}
Although Focus-DETR has designed a delicate token scoring mechanism and fine-grained feature enhancement methods, more hierarchical semantic grading strategies, such as object boundaries or centers, are still worth exploring.
In addition, our future work will be constructing a unified feature semantic scoring mechanism and fine-grained feature enhancement algorithm throughout the Transformer.

\section{Conclusion}
This paper proposes Focus-DETR to focus on more informative tokens for a better trade-off between computation efficiency and model accuracy. The core component of Focus-DETR is a multi-level discrimination strategy for feature semantics that utilizes a scoring mechanism considering both position and semantic information. Focus-DETR achieves a better trade-off between computation efficiency and model accuracy by precisely selecting foreground and fine-grained tokens for enhancement. Experimental results show that Focus-DETR has become the SOTA method in token pruning for DETR-like models. Our work is instructive for the design of transformer-based detectors.

{\small
\bibliographystyle{ieee_fullname}
\bibliography{egbib}
}

\clearpage

\appendix
\section{Appendix}
\subsection{More Implementation Details}
\label{A}

\subsubsection{Cascade Structure}
\label{sec:cascade}
In order to increase the fault tolerance of our model, we gradually reduce the scope of foreground regions through a cascade structure. As we show in Section~\textcolor{red}{3.4}, the computational complexity of deformable attention~\cite{Deformable_DETR} is linear with the number of preserved tokens. Therefore, there is no significant difference in complexity between the even structures (e.g., \{0.4,0.4,0.4,0.4,0.4,0.4\} and the cascade structures(e.g.,\{0.65,0.55,0.45,0.35,0.25,0.15\}). Table~\ref{tab:keep-ratio_supp} lists different average $keep-ratio$ and corresponding ratios of different layers designed in this paper. 
\begin{table}[htbp]
	\scriptsize
	\centering
	\setlength{\tabcolsep}{5mm}
	\begin{tabular}{c|c}
		\toprule
		Average $keep-ratio$ & Ratios \\
		\midrule
		0.1 & \{0.1,~0.1,~0.1,~0.1,~0.1,~0.1\}\\
		\midrule
		0.2 & \{0.3,~0.3,~0.2,~0.2,~0.1,~0.1\}\\
		\midrule
		0.3 & \{0.5,~0.4,~0.3,~0.3,~0.2,~0.1\}\\
		\midrule
		0.4 & \{0.65,0.55,0.45,0.35,0.25,0.15\}\\
		\midrule
		0.5 & \{0.75,0.65,0.55,0.45,0.35,0.25\}\\
		\bottomrule
	\end{tabular}%
	\caption{Detailed cascade keep-ratio desiged by Focus-DETR.}
	
	\label{tab:keep-ratio_supp}
\end{table}

\begin{figure}[t]
	\begin{center}
		\includegraphics[width=1.0\linewidth]{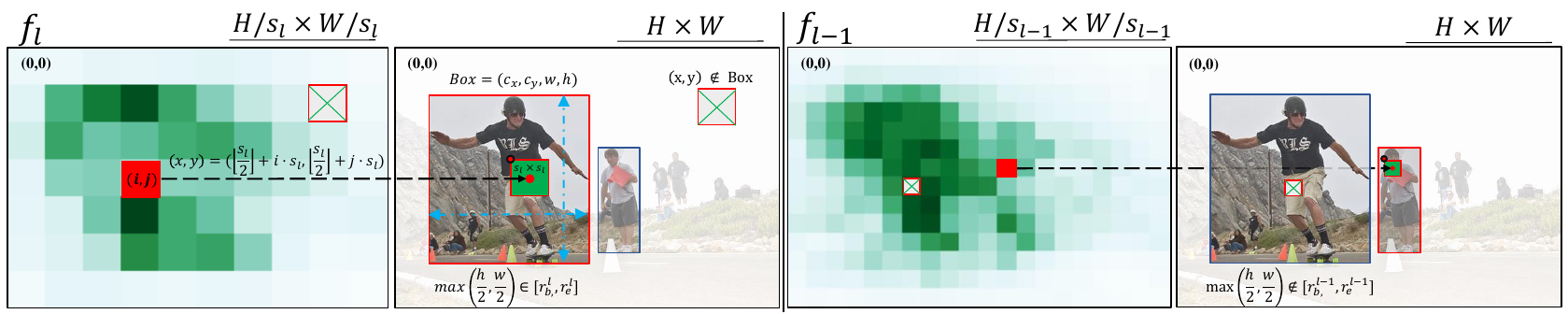}
	\end{center}
	\vspace{-0.7em}
	\caption{Visualization of the label assignment process. $f_l$, $f_{l-1}$ are feature maps with different scales~($s_l, s_{l-1}$).
	}
	\label{fig:appendix_1}
	\vspace{-1.6em}
\end{figure}

\subsubsection{Label Assignment}
\label{sec:overlap}

Unlike the traditional label assignment scheme for multi-scale feature maps, the ranges are allowed to overlap between the two adjacent feature scales to enhance the prediction near
the boundary. This strategy increases the number of foreground samples while ensuring that the multi-scale feature map predicts object heterogeneity. Intuitively, we assign the interval boundaries to be a series of integer power of two. As shown in Table~\ref{tab:overlap}, our overlapping interval setting improves the detection accuracy of the model when compared to non-overlapping ones using similar interval boundaries. As shown in Fig.~\ref{fig:appendix_1}, we present a visualization of the mapping between ground truth boxes in the original image and tokens from feature maps with different scales. 

\begin{table}[htbp]
	\scriptsize
	\centering
	\setlength{\tabcolsep}{0.7mm}
	\begin{tabular}{c|c|ccc}
		\toprule
		Model&Interval & $AP$& $AP_{50}$& $AP_{75}$   \\
		\midrule
		\multirow{2}[0]{*}{non-overlapping}&
		\{[-1, 64], [64, 128], [128, 256], [256, $\infty$]\} & 50.2&68.2&54.9  \\
		&\{[-1, 128], [128,256], [256,512], [512, $\infty$]\}&50.2& 68.1& 54.8 \\
		\midrule
		overlapping&\{[-1, 64], [64, 256], [128, 512], [256, $\infty$]\}&50.4& 68.5& 55.0\\
		\bottomrule
	\end{tabular}%
	\caption{Effect of preset scale intervals of multi-scale feature maps on experimental performance. Interval represents different scale intervals of multi-scale feature maps and $\infty=999999$ in experiments.}
	\label{tab:overlap}
\end{table}

\begin{table*}[htbp]
	\scriptsize
	\centering
	\setlength{\tabcolsep}{3.0mm}
		\begin{tabular}{c|c|c|cccccc|ccc}
			\toprule
			Model & Epochs &Backbone&$ AP$& $AP_{50}$& $AP_{75}$& $AP_{S}$& $AP_{M}$& $AP_{L}$     & Params& GFLOPs & FPS \\
			\midrule
			Deformable-DETR &Swin-T & 50 & 48.0& 68.0& 52.0& 30.3& 51.4 &63.7& 41M &185 & --\\
			\midrule
			Sparse DETR&Swin-T&50&49.1& 69.5& 53.5& 31.4& 52.5 &65.1&41M&129 &18.9\\
			\midrule
			\multirow{3}[0]{*}{{\textbf{Focus-DETR}}}&Swin-T& 36 & 52.5&{70.9}&{57.5}&{34.8}&{55.8}&{67.6}&{49M}&{163}&{15.3}     \\
			&Swin-B-224-22K&{36}&{56.0}&{74.8}&{61.1}&{40.1}&{59.5}&{72.0}&{109M}&{368}&{15.3}\\
			&Swin-B-384-22K&{36}&{56.2}&{75.1}&{61.7}&{38.2}&{60.0}&{72.5}&{109M}&{390}&{8.5}\\
			\bottomrule
		\end{tabular}%
	\caption{Results for our Focus-DETR using Swin Transformer as the backbone. Herein, Swin-T indicates the tiny version pretrained on ImageNet-1K~\cite{ImageNet}. Swin-B-224-22K represents the base version pretrained on ImageNet-22K~\cite{ImageNet} and the resolution of training set is 224. All reported FPS are measured on a NVIDIA V100 GPU. }
	\label{tab:swin}%
	\vspace{-1.0em}
\end{table*}%

\subsection{Supplementary Experiments} \label{B}

\subsubsection{Using Swin Transformer as the Backbone} \label{swin_backbone_ana}
When using Swin Transformer~\cite{SwinTransformer} as the backbone, Focus-DETR also achieves excellent performance. As shown in the following table, when Focus-DETR uses Swin-T as the backbone, the AP reaches 51.9 and achieve 56.0AP using Swin-B-224-22K and 55.9AP using Swin-B-384-22K. Compared with Deformable DETR~\cite{Deformable_DETR} and Sparse DETR~\cite{sparse_DETR}, our model achieves significant performance improvements, as shown in Table~\ref{tab:swin}. 

\subsubsection{Convergence Analysis}
In order to better observe the changes in model performance with the training epoch, we measured the changes in Focus-DETR test indicators and compared them with DINO. Experimental results show that Focus-DETR outperforms DINO even at 12 epochs when using ResNet50 as the backbone, as shown in Table~\ref{tab:compare_supp}. In addition, we found that the Focus-DETR reached the optimal training state at 24 epochs due to special foreground selection and fine-grained feature enhancement.

\subsubsection{Apply Dual Attention to Other Models}\label{sec:dual_attn_model}
As we mentioned in Section \textcolor{red}{4.3} of the main text, a precise
scoring mechanism is critical to the proposed dual attention. We add
the experiments of applying the encoder with dual attention
to those models equipped with Sparse DETR, such as Deformable DETR~\cite{Deformable_DETR}, DN DETR~\cite{DN_DETR} and DINO~\cite{DINO}. As shown in Table~\ref{tab:correlation_supp}, the proposed dual attention for fine-grained tokens enhancement brings only +0.3AP in Deformable DETR(two-stage), 0.0AP in Deformable DETR(without two-stage), -0.1AP in DN-Deformable-DETR and +0.3 AP in DINO. Results show us that untrusted fine-grained tokens do not bring significant performance	gains, which is still inefficient compared to Focus-DETR.
\begin{table}[htbp]
	\scriptsize
	\centering
	\setlength{\tabcolsep}{1.1mm}
	\begin{tabular}{c|c|c|c|c|c|c|c|c}
		\toprule
		Model                       & Backbone                & Epochs & $AP$   & $AP_{50}$ & $AP_{75}$ & $AP_S$  & $AP_M$  & $AP_L$  \\ \midrule
		\multirow{3}{*}{DINO~\cite{DINO}}       & \multirow{3}{*}{R-50}   & 12     & 49.0 & 66.6 & 53.5 & 32.0 & 52.3 & 63.0 \\ 
		&                         & 24     & 50.4 & 68.3 & 54.8 & 33.3 & 53.7 & 64.8 \\ 
		&                         & 36     & 50.9 & 69.0 & 55.3 & 34.6 & 54.1 & 64.6 \\ \midrule
		\multirow{9}{*}{\textbf{Focus-DETR}} & \multirow{3}{*}{R-50}   & 12     & 48.8 & 66.8 & 52.8 & 31.7 & 52.1 & 63.0 \\ 
		&                         & 24     & 50.3 & 68.4 & 55.1 & 33.9 & 53.5 & 64.4 \\ 
		&                         & 36     & 50.4 & 68.5 & 55.0 & 34.0 & 53.5 & 64.4 \\ \cline{2-9} 
		& \multirow{3}{*}{R-101}  & 12     & 50.8 & 69.3 & 55.5 & 32.8 & 54.5 & 65.8 \\ 
		&                         & 24     & 51.2 & 69.7 & 55.9 & 32.9 & 54.8 & 65.6 \\ 
		&                         & 36     & 51.4 & 70.0 & 55.6 & 34.2 & 55.0 & 65.5 \\ \cline{2-9} 
		& \multirow{3}{*}{Swin-T} & 12     & 49.9 & 68.2 & 54.3 & 32.9 & 52.8 & 65.1 \\ 
		&                         & 24     & 51.9 & 70.4 & 56.6 & 35.4 & 54.9 & 67.0 \\ 
		&                         & 36     & 52.5 & 70.9 & 57.5 & 34.8 & 55.8 & 67.6 \\ \bottomrule
	\end{tabular}
	\caption{Focus-DETR uses different backbones at different training epochs and provides comparison results with DINO~\cite{DINO}. R-50 and R-101 is ResNet backbone, Swin-T represents Swin Transformer of the tiny version.}
	\label{tab:compare_supp}%
\end{table}
\begin{table}[htbp]
	\scriptsize
	\centering
	\setlength{\tabcolsep}{1.8mm}
	\begin{tabular}{c|c|c|c|c}
		\toprule
		Model&epoch&$AP$ &GFLOPs&FPS  \\
		\hline
		Deformable DETR~\textcolor{blue}{(priori)}&50& 46.2 &177&19 \\
		+ Sparse DETR~($\alpha=0.3$)&50 &46.0&121&23.2 \\
		+ Sparse DETR(dual attention)~($\alpha=0.3$)&50 &46.3&123&23.0 \\
		or +\textbf{ Focus-DETR}~($\alpha=0.3$)&50&46.6&123& 23.0\\
		\hline
		Deformable DETR~\textcolor{blue}{(learnable)}&50&45.4&173&19 \\
		+ Sparse DETR~($\alpha=0.3$)& 50&43.5&118&24.2 \\
		+ Sparse DETR(dual attention)~($\alpha=0.3$)& 50&43.5&120&23.9 \\
		or + \textbf{Focus-DETR}~($\alpha=0.3$)&50&45.2&120&23.9 \\
		\hline
		DN-Deformable-DETR~\textcolor{blue}{(learnable)}&50&48.6&195&18.5 \\
		+ Sparse DETR~($\alpha=0.3$)&50&47.4&137&23.9 \\
		+ Sparse DETR(dual attention)~($\alpha=0.3$)&50 &47.3&138&23.7 \\
		or + \textbf{Focus-DETR}~($\alpha=0.3$)&50&48.5&138&23.6 \\
		\hline
		DINO~\textcolor{blue}{(mixed)}&36 &50.9&279&14.2 \\
		+ Sparse DETR~($\alpha=0.3$)&36 &48.2&152&20.2 \\
		+ Sparse DETR(dual attention)~($\alpha=0.3$)&36&48.5&154&20.0 \\
		or + \textbf{Focus-DETR}~($\alpha=0.3$)&36 &50.4&154&20.0 \\
		\bottomrule
	\end{tabular}%
	
	\caption{Apply dual attention to the classic models equipped with Sparse DETR and compare them with Focus-DETR.}
	
	\small \label{tab:correlation_supp}%
	\vspace{-1.7em}
\end{table}

\subsection{Visualization}
\begin{figure*}
	\begin{center}
		\includegraphics[width=15cm]{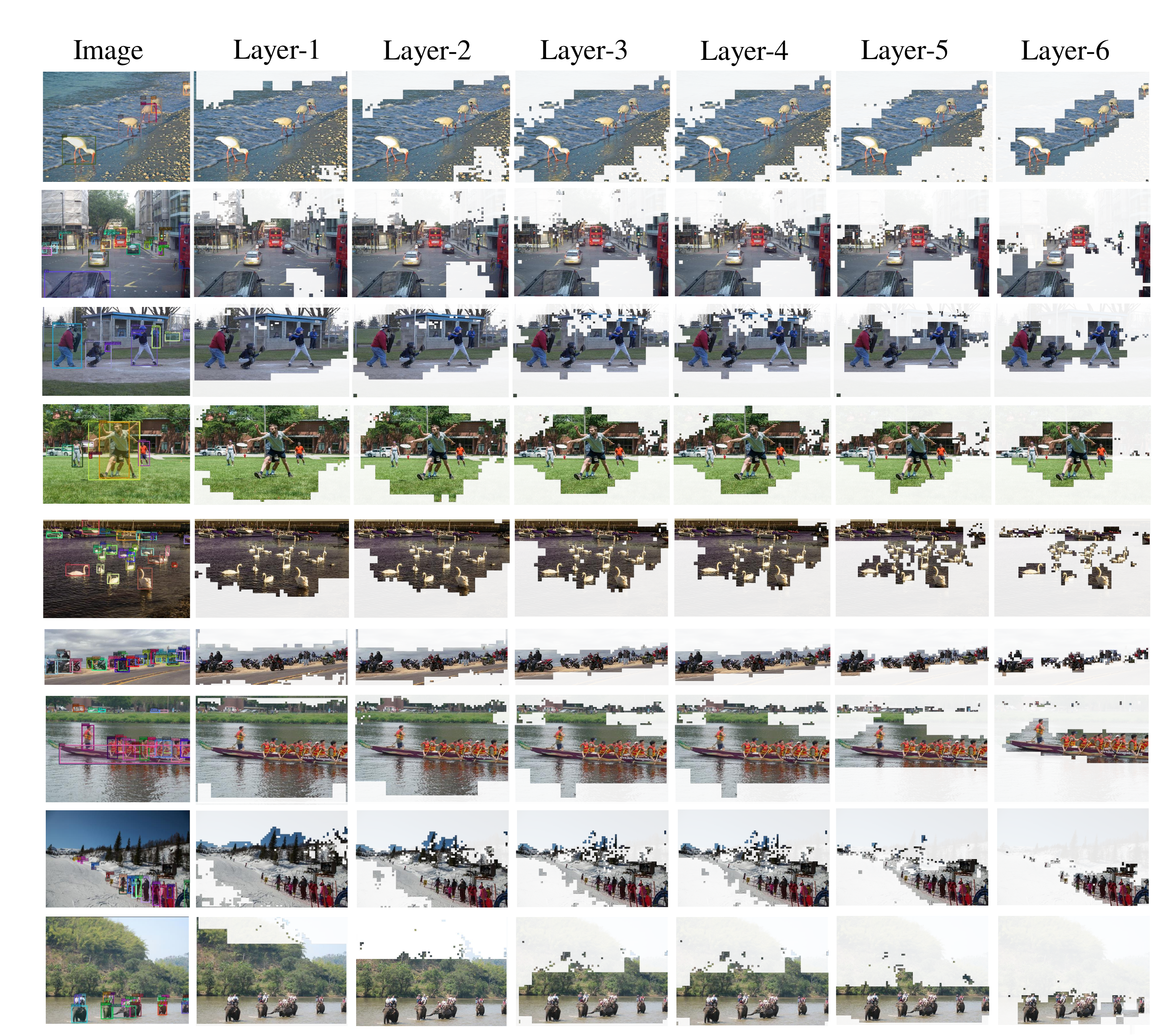}
	\end{center}
	\caption{Visualization result of foreground tokens reserved at each encoder layer, and final detection results are provided. Layer-$\{1,2,3,...\}$ indicates different encoder layers.
	}
	\label{fig:vision3_supp}
\end{figure*}

As shown in Fig.~\ref{fig:vision3_supp}, we visualize eight test images with diverse categories, complex backgrounds, overlapping targets, and different scales. We analyze the foreground features retained by different encoder layers. Visualization results show that foreground areas focus on a more refined area layer by layer in the encoder. Specifically, the result of Layer-6  captures a more accurate foreground with fewer tokens. The final test results of Focus-DETR are also presented, as shown in the first column.

\begin{figure*}
	\begin{center}
		\includegraphics[width=16cm]{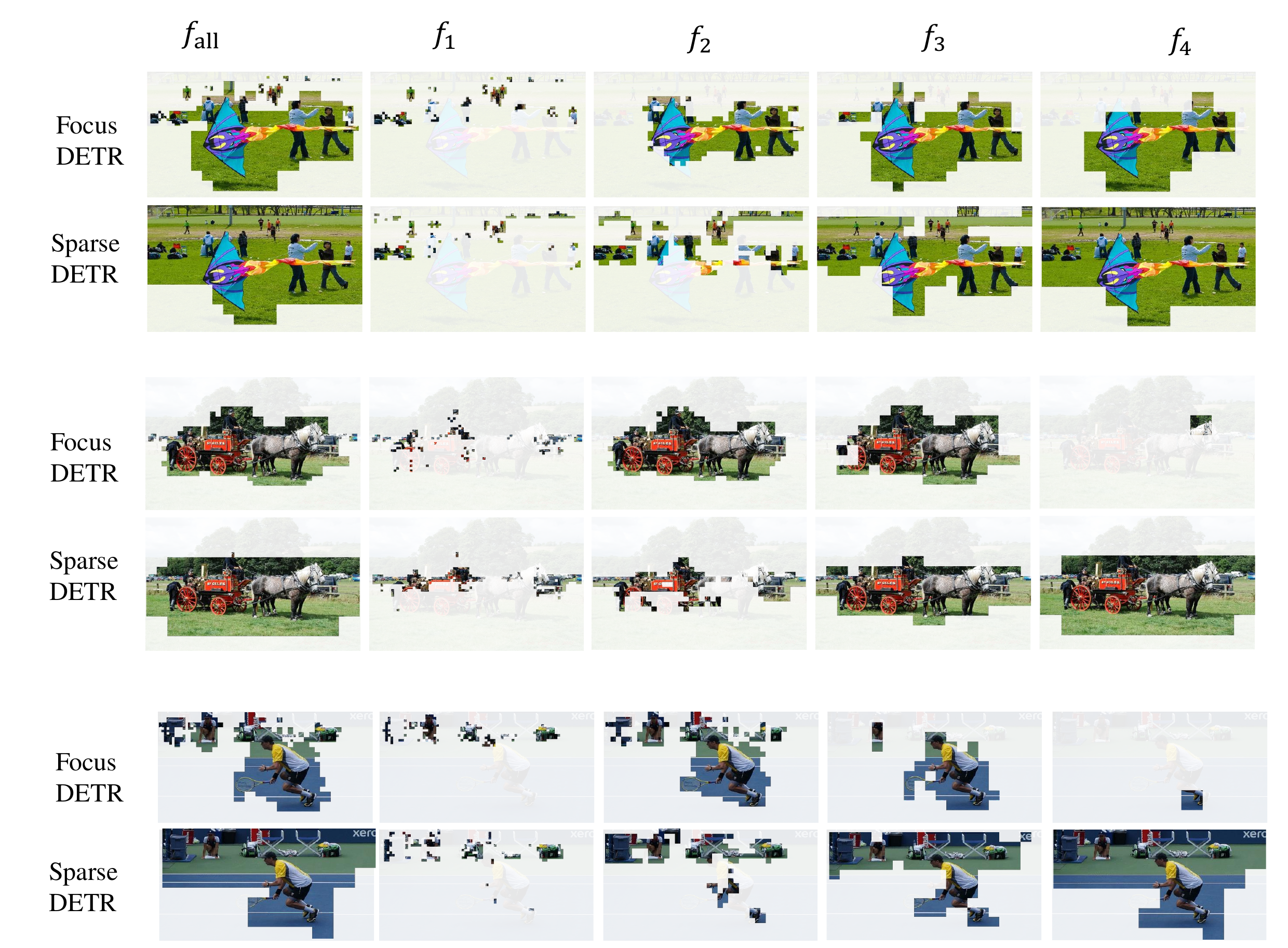}
	\end{center}
	\caption{Visualized comparison result of foreground tokens reserved in different feature maps. We analyze the difference between Focus-DETR and Sparse DETR~\cite{sparse_DETR} by using three images with obvious object scale differences. $f_{all}$ is the tokens retained by all feature maps, $\{f_1,~f_2,~f_3,~f_4\}$ represents different feature maps.
	}
	\label{fig:vision5_supp}
\end{figure*}

In addition, we compare the differences of multi-scale feature maps retention object tokens due to our label assignment strategy. We also visualize Sparse DETR~\cite{sparse_DETR} to demonstrate the performance. As shown in first column of Fig.~\ref{fig:vision5_supp}, Focus-DETR can obtain more precise foreground than Sparse DETR. According to the results of $\{f_1,~f_2,~f_3,~f_4\}$, the multi-scale feature map of Focus-DETR can retain tokens according to different object scales, which further proves the advantages of our tag allocation and top-down score modulations strategy.

\label{C}

\end{document}